\documentclass[11pt,a4paper]{article}
\usepackage[utf8]{inputenc}
\usepackage[nohyperref]{resources/naaclhlt2018}
\usepackage[hyperindex,breaklinks]{hyperref}
\usepackage{times}
\usepackage{latexsym}
\usepackage{amssymb}
\usepackage[colorinlistoftodos,prependcaption,textsize=tiny]{todonotes}
\usepackage{siunitx}
\usepackage{rotating}
\usepackage{graphicx}

\usepackage{xspace}
\usepackage{url}
\usepackage{cleveref}
\usepackage{booktabs}

\usetikzlibrary{decorations.pathreplacing,calc,fit,shapes.misc,shadows,arrows,decorations.pathmorphing}

\newcommand{\clstm}{C-LSTM\xspace}

\newcommand{\ccm}{CC\xspace}
\newcommand{\scm}{arXiv\xspace}

\newcommand{\scem}{\scm}
\newcommand{\ccem}{\ccm}

\newcommand{\furl}[1]{\footnote{\url{#1}}}

\aclfinalcopy 

\title{ClaiRE at SemEval-2018 Task 7 - Extended Version}

\author{Lena Hettinger, Alexander Dallmann, Albin Zehe, Thomas Niebler and Andreas Hotho \\
DMIR Group \\
University of Wuerzburg\\
{\tt \{hettinger,dallmann,zehe,niebler,hotho\}}\\
{\tt@informatik.uni-wuerzburg.de} \\}

\date{}

\begin{document}
    \maketitle
    \begin{abstract}
    In this paper we describe our post-evaluation results for SemEval-2018 Task 7 on classification of semantic relations in scientific
    literature for clean (subtask 1.1) and noisy data (subtask 1.2).
    This is an extended version of our workshop paper~\cite{hettinger2018semeval} including further technical
    details (Sections~\ref{subsec:clstm} and~\ref{subsec:clstm_results})
    and changes made to the preprocessing step in the post-evaluation phase (Section~\ref{subsec:prepro}).
    Due to these changes \textbf{Cla}ss\textbf{i}fication of \textbf{R}elations using \textbf{E}mbeddings (ClaiRE)
    achieved an improved F1 score of \SI{75.11}{\percent} for the first subtask and \SI{81.44}{\percent} for the
    second.
\end{abstract}

    \section{Introduction}
\label{sec:intro}
The goal of SemEval-2018 Task 7 is to extract and classify semantic relations
between entities into six categories that are specific to scientific
literature \citep{SemEval2018Task7}.
In this work, we focus on the subtask of classifying relations between entities
in manually (subtask 1.1) and automatically annotated and therefore noisy data (subtask 1.2).
Given a pair of related entities, the task is to classify the type
of their relation among the following options:
\texttt{Compare},
\texttt{Model-Feature},
\texttt{Part\_Whole},
\texttt{Result},
\texttt{Topic} or
\texttt{Usage}.
Relation types are explained in detail in the task description paper~\cite{SemEval2018Task7}.
The following sentence shows an example of a \texttt{Result} relation between the two entities
\textbf{combination methods}
and
\textbf{system performance}:
\begin{quote}
\textbf{Combination methods}
are an effective way of improving
\textbf{system performance}.
\end{quote}

This sentence is a good example for two challenges we face in this task.
First, almost half of all entities consist of noun phrases which has to be considered when constructing features.
Secondly, the vocabulary is domain dependent and therefore background knowledge should be adopted.

Previous approaches for semantic relation classification tasks mainly employed two strategies.
Either they made use of a lot of hand-crafted features or they utilized a neural network with as few background
knowledge as possible.
The winning system of an earlier SemEval challenge on relation classification \citep{hendrickx2009semeval} adopted the
first approach and achieved an F1 score of 82.2\% \citep{rink2010utd}.
Later, other works outperformed this approach by using CNNs with and without hand-crafted features
\citep{santos2015classifying, xu2015semantic} as well as RNNs \citep{miwa2016endtoend}.

\paragraph{Approach} We present two approaches that use different levels of preliminary information.
Our first approach is inspired by the winning method of the SemEval-2010 challenge \citep{rink2010utd}.
It models semantic relations by describing the two \emph{entities},
between which the semantic relation holds,
as well as the words between those entities.
We call those in-between words the \emph{context} of the semantic relation.
We classify relations by using an SVM on lexical features,
such as part-of-speech tags.
Additionally we make use of semantic background knowledge and add pre-trained word embeddings to the SVM, as word
embeddings have been shown to improve performance in a series of NLP tasks,
such as sentiment analysis \citep{kim2014convolutional},
question answering~\cite{chen2017reading}
or
relation extraction~\cite{dligach2017neural}.
Besides using existing word embeddings generated from a general corpus, we also train embeddings on scientific articles that better reflect scientific vocabulary.

In contrast, our second approach relies on word embeddings only,
which are fed into a convolutional long-short term memory (\clstm)
network, a model that combines convolutional and recurrent neural networks~\cite{zhou2015clstm}.
Therefore no hand-crafted features are used.
Because both CNN and RNN models have shown good performance for this task,
we assume that a combination of them will positively impact classification
performance compared to the individual models.

\paragraph{Results}
By combining Lexical information and domain-Adapted Scientific word Embeddings,
our system ClaiRE originally achieved an F1 score of 74.89\% for the first subtask with
manually annotated data and 78.39\% for the second subtask with automatically annotated data~\cite{hettinger2018semeval}.
Improving our preprocessing lifted this performance to 75.11\% and 81.44\% respectively.
Our results make a strong case for domain-specific word embeddings, as using
those improved our score by close to 5\%.

\paragraph{Paper Structure}
In \Cref{sec:features}, we describe the features that we used to characterize
semantic relations.
\Cref{sec:methods} shows how we classify the relation using an SVM and a C-LSTM neural network.
\Cref{sec:eval} presents the results, which are discussed in \Cref{sec:discussion}.
Finally, \Cref{sec:conclusion} concludes this work.

    \section{Features}\label{sec:features}

\begin{table*}[h!]
    \centering
    \small
    \begin{tabular}{ l l}
         \multicolumn{2}{l}{{\bfseries Example Sentence:} \textbf{Combination methods} are an effective way of improving
         \textbf{system performance}.}
         \\
         \toprule
         Lexical Feature Set	& Exemplary Boolean Features \\
         \midrule
         BagOfWords ($bow$)	& an, be, effective, improve, of, way \\
         POS tags ($pos$)	    & ADJ, ADP, DET, NOUN, VERB \\
         POS path ($pospath$)	    & VDANAV \\
         Distance ($dist$)		& 6 \\
         Levin classes ($lc$)		& 45 \\
         \midrule
         Entities
         without order ($ents$)	& combination methods, methods, system performance, performance \\
         Start entity ($startEnt$) & combination methods, methods \\
         End entitiy ($endEnt$)	& system performance, performance \\
         Similarity ($sim100$)		& 0.43 \\
         Similarity bucket ($simb$)  & q50 \\
        \bottomrule
    \end{tabular}
    \caption{Examples for lexical context and entity features.}
    \label{table:example}
\end{table*}

In this section, we describe the features which are used in our two approaches.
All sentences are first preprocessed before constructing boolean lexical features on the one hand and word embedding
vectors on the other.
Both feature groups are based on the entities of relations as well as the context in which those entities appear.

Apart from the \texttt{Compare} relation, all relation types are asymmetric, and therefore the distinction between
start and end entity of a relation is important.
If entities appear in reverse order, that means the end entity of a relation appears first in the sentence, this is
marked by a \emph{direction} feature which is part of the data set.

In our entrance example, \textbf{combination methods} denotes the start entity,
\textbf{system performance} the end entity,
and \textbf{are an effective way of improving} the context.

\subsection{Preprocessing}\label{subsec:prepro}
Early experiments showed that it is beneficial to filter the vocabulary of our data and reduce noise by leaving out
infrequent context words.
The best setting was found to be a frequency threshold of \num{5} on lemmatized words.
Therefore we discard a context word if its lemma appears less than \num{5} times in the corpus of the respective subtask.

\paragraph{Post-Evaluation changes}
Lemmas as well as POS tags were extracted with the help of SpaCy.\furl{https://spacy.io/}
We started and finished the challenge with version \texttt{2.0.2} and afterwards updated to version \texttt{2.0.9}.
This version update lead to a change of POS tags, with which our results improved.
During post-evaluation we also noticed an error in the preprocessing that caused two feature sets ($bow$ and $pos$) to
intermix.
Both the lemmas of pronouns as well as the POS-tags of pronouns were mapped to the same symbol 'PRON',
therefore we had to explicitly separate these two sets.
After resolving this intersection our results improved further.

\subsection{Context features}\label{subsec:context}
First we will explain feature construction based on the context of a relation.
Abbreviations for feature names are denoted in brackets.
Context is defined as the words between two entities.
Early tests showed that using those words described the relation better than
the words surrounding the relation entities.

\paragraph{Lexical}
We construct several lexical boolean features which are illustrated in
\Cref{table:example}.
First we apply a bag of words ($bow$) approach where each lemmatized word forms one boolean feature,
which for example takes 1 as value if the lemma \emph{improve} is present and 0 if it is not.
Second we determine whether the context words contain certain part-of-speech (POS) tags ($pos$), such as \emph{VERB}.
To represent the structure of the context phrase we add a path of POS tags feature, which contains the order in
which POS tags appear ($pospath$).
The distance feature depicts whether the POS-path and therefore the context phrase has a certain length ($dist$).

Additionally we add background knowledge by extracting the top-level Levin classes of intermediary verbs from
VerbNet\furl{http://verbs.colorado.edu/~mpalmer/projects/verbnet.html} ($lc$), a verb lexicon compatible with WordNet.
It contains explicitly stated syntactic and semantic information, using
Levin verb classes to systematically construct lexical entries~\cite{schuler2005verbnet}.
For example the verb \emph{improve} belongs to class 45.4, which is described by Levin as consisting of
``alternating change of state`` verbs.\furl{http://www-personal.umich.edu/~jlawler/levin.verbs}

\paragraph{Embeddings} Aside from lexical features we also use word embedding vectors to leverage information
from the context of entities ($c$).
For each filtered context word we extract its word embeddding from a pre-trained corpus, where out-of-vocabulary words
(OOV) are represented by the zero vector.
The individual word vectors are later applied to train a \clstm.

In contrast, for use in an SVM we found it beneficial to represent the
context embedding features as the average over all context word embeddings.

\subsection{Entity features}\label{subsec:entities}
In the second set of features, we model the relation entities themselves as they may be connected to a certain
relation class.
For example, the token \emph{performance} or one form of it mostly appears as an end entity of a \texttt{Result}
relation, and in the rare case when it represents a start entity, it is almost always part of a \texttt{Compare}
relation.
Therefore we leverage information about entity position for the creation of lexical and embedding entity features.

\paragraph{Lexical} For the creation of boolean lexical features, we first take the lowercased string of each entity
and construct up to three distinct features from it.
One feature which marks its general appearance in the corpus without order ($ents$) and one each if it occurs as
start ($startEnt$) or end ($endEnt$) entity of a relation, taking its direction into account.
Additionally we add the head noun to the respective feature set if the entity consists of a nominal phrase to create
greater overlap between instances.

Furthermore we measure the semantic similarity of the relation entities using the cosine of the corresponding
word embedding vectors ($sim100$).
While the cosine takes every value from [-1, 1] in theory, we cut off after two digits to reduce the feature space and
get 99 boolean similarity features for our corpus.
To again enable learning across instances we additionally discretize the similarity range and form another five
boolean similarity features ($simb$) that capture into which of the following buckets the similarity score falls:
$q0=[-1, 0), q25=[0, 0.25), q50=[0.25, 0.5), q75=[0.5, 0.75), q100=[0.75, 1]$ (values below zero are very rare in this corpus).

\paragraph{Embeddings} Similar to the context features we also want to add word embeddings of entities to our
entity feature set.
This is not straighforward as more than 44\% of all entities consist of nominal phrases, while a word embedding usually
corresponds to a single word.
By way of comparison, the proportion of nominals in the relation classification corpus of the SemEval-2010 challenge
was only 5\%.
Thus we tested different strategies to obtain a word embedding for nominal phrases and found that averaging over the
individual word vectors of the phrase yielded the best results for this task.
These word embeddings for start ($e_s$) and end ($e_e$) entities of relations were then presented to our two classification
methods, which will be described in detail in the following section.

    \section{Classification Methods}\label{sec:methods}
We utilize two different models for classifying semantic relations:
an SVM which incorporates both the lexical and embedding features described in \Cref{sec:features}
and a Convolutional Long Short Term Memory (C-LSTM) neural network that only uses word embedding vectors

\subsection{SVM}\label{subsec:svm}

\begin{figure*}
\centering
	\begin{tikzpicture}
	\tikzset{every node/.style={node distance=0.2cm,inner sep=1mm,font=\footnotesize}}
	\tikzset{featureset/.style={shape=rounded rectangle, draw, minimum height=2em,font=\footnotesize, minimum width=2.5em,inner sep=1mm}}
	\tikzset{doc/.style={draw, minimum height=4em, minimum width=3em, 
		  fill=white, 
		  double copy shadow={shadow xshift=4pt, 
				 shadow yshift=4pt, fill=white, draw}}}
	
	
	\node[featureset] (f11) {$bow$};
	\node[featureset, right=of f11] (f12) {$pos$};
	\node[featureset, right=of f12] (f13) {$pospath$};
	\node[featureset, right=of f13] (f14) {$dist$};
	\node[featureset, right=of f14] (f15) {$lc$};
	
	\node[below=0.5cm of f11] (f11size) {\num{1129}};
	\node[below=0.5cm of f12] (f12size) {\num{13}};
	\node[below=0.5cm of f13] (f13size) {\num{965}};
	\node[below=0.5cm of f14] (f14size) {\num{23}};
	\node[below=0.5cm of f15] (f15size) {\num{44}};
	
	\draw[->,dashed] (f11) -- (f11size);
	\draw[->,dashed] (f12) -- (f12size);
	\draw[->,dashed] (f13) -- (f13size);
	\draw[->,dashed] (f14) -- (f14size);
	\draw[->,dashed] (f15) -- (f15size);

	\node[featureset, right=of f15] (f31) {$c$};
	\node[featureset, right=of f31] (f32) {$e_s$};
	\node[featureset, right=of f32] (f33) {$e_e$};

	\node[below=0.5cm of f31] (f31size) {\num{300}};
	\node[below=0.5cm of f32] (f32size) {\num{300}};
	\node[below=0.5cm of f33] (f33size) {\num{300}};
	\draw[->,dashed] (f31) -- (f31size);
	\draw[->,dashed] (f32) -- (f32size);
	\draw[->,dashed] (f33) -- (f33size);
	

	\node[featureset, right=of f33] (f21) {$ents$};
	\node[featureset, right=of f21] (f22) {$startEnt$};
	\node[featureset, right=of f22] (f23) {$endEnt$};
	\node[featureset, right=of f23] (f24) {$sim100$};
	\node[featureset, right=of f24] (f25) {$simb$};
	\node[featureset, right=of f25] (f26) {$dir$};

	\node[below=0.5cm of f21] (f21size) {\num{3097}};
	\node[below=0.5cm of f22] (f22size) {\num{1831}};
	\node[below=0.5cm of f23] (f23size) {\num{1783}};
	\node[below=0.5cm of f24] (f24size) {\num{99}};
	\node[below=0.5cm of f25] (f25size) {\num{5}};
	\node[below=0.5cm of f26] (f26size) {\num{1}};

	\draw[->,dashed] (f21) -- (f21size);
	\draw[->,dashed] (f22) -- (f22size);
	\draw[->,dashed] (f23) -- (f23size);
	\draw[->,dashed] (f24) -- (f24size);
	\draw[->,dashed] (f25) -- (f25size);
	\draw[->,dashed] (f26) -- (f26size);

	\node[draw,fit=(f11) (f26) (f11size) (f12size),inner sep=2mm] (box) {};

	\draw[decorate,decoration={brace,mirror}] (f11.west |- box.south) + (0,-0.1) -- ($(f15.east |- box.south) + (0,-0.1)$) node[midway,label=below:{\footnotesize \num{2174} lexical context}] (f1size) {};

	\draw[decorate,decoration={brace,mirror}] (f21.west |- box.south) + (0,-0.1) -- ($(f26.east |- box.south) + (0,-0.1)$) node[midway,label=below:{\footnotesize 6816 lexical entity}] (f2size) {};
	
	\draw[decorate,decoration={brace,mirror}] (f31.west |- box.south) + (0,-0.1) -- ($(f33.east |- box.south) + (0,-0.1)$) node[midway,label=below:{\footnotesize 900 embedding}] (f3size) {};
	
	\draw[decorate,decoration={brace}] (f11.west |- box.north) + (0,0.1) -- ($(f31.east |- box.north) + (0,0.1)$) node[midway,label=above:{\footnotesize context}] () {};
	
	\draw[decorate,decoration={brace}] (f32.west |- box.north) + (0,0.1) -- ($(f26.east |- box.north) + (0,0.1)$) node[midway,label=above:{\footnotesize entity}] () {};

	\draw[decorate,decoration={brace,mirror}] (f11.west |- f1size.south) + (0,-0.5) -- ($(f26.east |-  f2size.south) + (0,-0.5)$) node[midway,label=below:{\footnotesize 9886 features}] (overallsize) {};
	
	\draw ($(f15.east)!0.5!(f21.west)$ |- box.north) -- ($(f15.east)!0.5!(f21.west)$ |- box.south);
	
	\draw[dotted] let \p1 = (box.north), \p2 = ($(f15.east)!0.5!(f31.west)$), \p3 = (box.south) in (\x2,\y1) -- (\x2,\y3);
	\draw[dotted] let \p1 = (box.north), \p2 = ($(f33.east)!0.5!(f21.west)$), \p3 = (box.south) in (\x2,\y1) -- (\x2,\y3);
	\draw[] let \p1 = (box.north), \p2 = ($(f31.east)!0.5!(f32.west)$), \p3 = (box.south) in (\x2,\y1) -- (\x2,\y3);
	
	\node[fit={(box) ($(overallsize)+(0,-1em)$)}] (features) {};

\end{tikzpicture}
	\caption{Feature vector used in the SVM. Numbers hold true for subtask 1.1, including 1.2 data}
	\label{fig:vector}
\end{figure*}
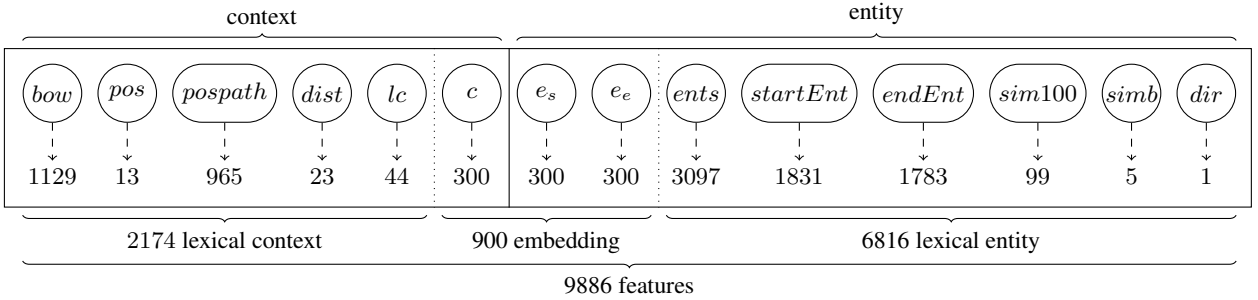

To fully exploit our hand-crafted lexical features we employ a traditional classifier.
In comparison to Naive Bayes, Decision Trees and Random Forests we found a Support Vector Machine to perform best for
this task.
Instead of utilizing the decision function of the SVM to predict test labels we decided to make use of the probability
estimates according to \citet{PairwiseCoupling-Estimates} as this proved to be more successful.
As mentioned before, the lexical features are fed into the SVM as boolean features whereas the word embeddings are
normalized using MinMax-Scaling to the range $[0,1]$ to make it easier for the SVM to handle both feature groups
(Fig.~\ref{fig:vector}).

\subsection{C-LSTM}\label{subsec:clstm}
In contrast to SVM, neural network models do not necessarily rely on handcrafted features and are therefore faster to implement.
We experiment with \clstm\cite{zhou2015clstm} which extracts a sentence representation by combining one-dimensional
convolution and an LSTM network and uses the representation to perform a classification.

\begin{figure}[h!]
	\centering
	\includegraphics{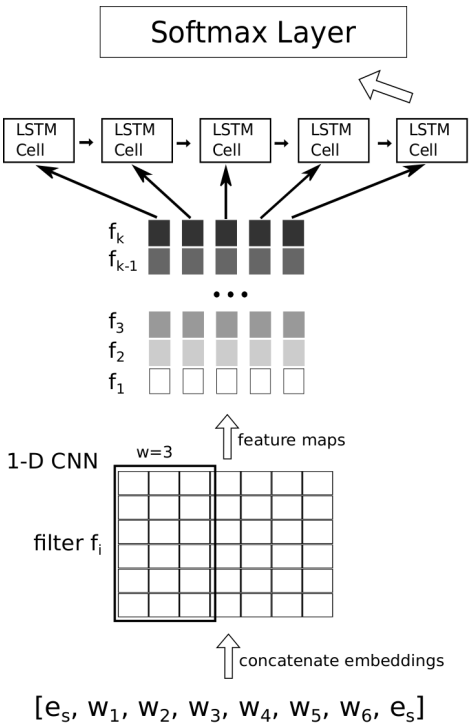}
	\caption{An illustration of the model architecture of \clstm.}
    \label{fig:clstm}
\end{figure}

\clstm extracts a sentence representation in the following steps.
First embeddings for all words $w_i \in \mathbb{R}^{v}$ are obtained from a pre-computed embedding table $E \in
\mathbb{R}^{v \times |V|}$ where $v$ is the embedding size and $|V|$ denotes the size of the vocabulary.
For entities that are nominal phrases the average over the individual word embeddings is used.
This results in a sequence of embedding vectors $s = \left[e_s, w_1, w_2, \cdots, w_n, e_e\right]$ of length $l_s$ where
$e_1, e_2 \in \mathbb{R}^{v}$ are embeddings representing entities and the $w_i$ represent the context word embeddings.
Next the embedding vectors in $s$ are concatenated to form an input matrix $I \in \mathbb{R}^{v \times l_s}$ for the
one-dimensional convolution layer.
For computational reasons a matrix $\hat{I} \in \mathbb{R}^{v \times l_{max}}$ is obtained by right padding $I$ with a zero
token to the maximum sequence length $l_{max}$ in the corpus.
After that $k$ feature maps $f_i \in \mathbb{R}^{m}$ with $m$ being the number of features in each map are computed over
$\hat{I}$ using a one-dimensional convolution layer with $k$ filters of window size $ws$ and stride $st$.
The resulting feature map matrix $C \in \mathbb{R}^{k \times m}$ is then split along the second axis into a sequence $c$ with
individual elements $c_i \in \mathbb{R}^{k}$ and length $l_{c}=m$.
Finally $c$ is used as input to an LSTM network with the last output being a representation of the input sequence.
A softmax layer is used to compute label scores from the sentence representation.
See \Cref{fig:clstm} for an illustration of the model.

    \section{Evaluation}\label{sec:eval}
After describing the two models we employ for relation classification, we now portray the data set we use and present
results for both SVM and \clstm in detail.
Results are reported as micro-F1 and macro-F1.
The latter is the official evaluation score of the SemEval Challenge.
We describe the experimental setup for both models and compare different feature sets and pre-trained embeddings.

\subsection{Data and Background Knowledge}
\begin{table}
    \centering
    \small
        \begin{tabular}{ l  r  r  r}
            \toprule label
            		  & subtask 1.1
            		  & subtask 1.2
            		  & total\\
            \midrule
            COMPARE		& 95 (\phantom{0}8\%) 		& 41 (\phantom{0}3\%)		& 136 (\phantom{0}5\%)\\
            MODEL-F.	& 326 (27\%)	& 174 (14\%)	& 500 (20\%)\\
            PART\_W.	& 234 (19\%)	& 196 (16\%)	&  430 (17\%)\\
            RESULT		& 72 (\phantom{0}6\%)		& 123 (10\%)		& 195 (\phantom{0}8\%)\\
            TOPIC		& 18 (\phantom{0}1\%)		& 243 (20\%)	& 261 (11\%)\\
            USAGE		& 483 (39\%)	& 468 (38\%)	& 951 (38\%)\\
            \bottomrule
        \end{tabular}
    \caption{\label{class-dis} Distribution of class labels for training data as absolute and relative values.}
\end{table}

We evaluate our approach on a set of scientific abstracts, $D_{test}$.
It consists of 355 semantic relations for each subtask which are similarly distributed as its respective training data set.
As training data we received \num{350} abstracts of scientific articles per subtask,
which resulted in \num{1228} labeled training relations for subtask 1.1 and \num{1245}
training instances for subtask 1.2 (c.f. \Cref{class-dis}).
We combine data sets of both subtasks for training, resulting in \num{2473} training examples in total ($D_{train}$).

\paragraph{Background Knowledge}
In our experiments, we compare different pre-trained word embeddings as a source of background knowledge.
As a baseline, we employ a publicly available set of 300-dimensional word embeddings trained with
GloVe~\cite{pennington2014glove} on the Common Crawl data\furl{http://commoncrawl.org/} (\textit{\ccem}).
To better reflect the semantics of scientific language, we trained our own scientific embeddings using
word2vec~\cite{mikolov2013distributed} on a large corpus of papers collected from arXiv.org\furl{https://arxiv.org}
(\textit{\scem}).

In order to create the scientific embeddings, we downloaded \LaTeX~sources for all papers
published in 2016 on arXiv.org using the provided dumps.\furl{https://arxiv.org/help/bulk_data}
After originally trying to extract the plain text from the sources, we found that it was more
feasible to first compile the sources to pdf (excluding all graphics etc.) and then use
pdftotext\furl{https://poppler.freedesktop.org} to convert the documents to plain text.
This resulted in a dataset of about \num{166000} papers.
Using gensim~\cite{rehurek_lrec}, for each document we extracted tokens of minimum length 1 with the wikicorpus
tokenizer and used word2vec to train 300-dimensional word embeddings on the data.
We kept most hyper-parameters at their default values, but limited the vocabulary to words
occurring at least 100 times in the dataset, reducing for example the noise introduced by
artifacts from equations.

\subsection{SVM}

\begin{table}
    \centering
    \small
    \begin{tabular}{ l c c c c}
        \toprule
         & \multicolumn{2}{c}{context}  &  \multicolumn{2}{c}{+ entities}
         \\
         data & macro F1 & micro F1 & macro F1 & micro F1 \\
         \midrule
        1.1 & 45.10 & 59.15 & 48.96 & 65.35 \\
        +1.2 & 46.95 & 61.97 & 66.03 & 70.14 \\
        \midrule
        CC & 51.14 & 64.79 & 70.31 & 73.24 \\
        arXiv & 51.55 & 64.79 & \textbf{75.11} & \textbf{77.46} \\
        \bottomrule
    \end{tabular}
    \caption{SVM results for subtask 1.1.}
    \label{table:subtask1}
\end{table}

After an extensive grid search per cross validation the best parameters for the SVM were found to be a rbf-kernel with
$C=100$ and $\gamma=0.001$ for both tasks.

Our post-evaluation results of the SVM for subtask 1.1. are shown in \Cref{table:subtask1}.
Adding entity features proves to be very beneficial compared to using only context features, as we
could improve macro-F1 by 16 points on average.
Results are further improved by enlarging the data set with the training samples of subtask 1.2 and by adding word
embeddings to the feature set.
While adding the \ccm embeddings enhances the micro-F1 by more than 4 points, our domain-adapted \scm embeddings
prove to perform even better and deliver the best result with a macro-F1 score of \SI{75.11}{\percent} and a
micro-F1 of \SI{77.46}{\percent}.

\begin{table}
    \centering
    \small
    \begin{tabular}{ l c c c c}
        \toprule
         & \multicolumn{2}{c}{context}  &  \multicolumn{2}{c}{+ entities}
         \\
         data & macro F1 & micro F1 & macro F1 & micro F1 \\
         \midrule
        1.2 & 68.61 & 71.27 & 73.49 & 81.41 \\
        +1.1 & 61.09 & 69.01 & 78.63 & 83.66 \\
        \midrule
        CC & 62.74 & 70.42 & 76.80 & \textbf{85.63} \\
        arXiv & 63.29 & 70.99 & \textbf{81.44} & 85.07 \\
        \bottomrule
    \end{tabular}
    \caption{SVM results for subtask 1.2.}
    \label{table:subtask2}
\end{table}

Similar observations can be made for subtask 1.2., as is pictured in \Cref{table:subtask2}.
Originally we achieved a micro-F1 score 74.89\% for the first subtask and 78.39\% for the second but adding the
changes noted in Section~\ref{subsec:prepro} led to an improvement of \SI{2}{\percent} on average~\cite{hettinger2018semeval}.

\subsection{C-LSTM}\label{subsec:clstm_results}
We fix the batch size and number of epochs to \num{128} and \num{100} respectively for all trained models.
Words are encoded using either \scm or \ccm embeddings.
The embeddings are not further optimized during training.
Cross-entropy is used as the loss function and the model is optimized using Adam~\cite{kingma2014adam} with the initial
learning rate set to $lr=0.002$, $\beta_1=0.9$, $\beta_2=0.999$, $\varepsilon=10^{-8}$.

To find the optimal hyperparameter configuration, we perform a random search~\cite{Bergstra2012RandomSF} on the
hyper-parameters \emph{number of filters}, \emph{filter width},
\emph{rnn cell units}, \emph{dropout rate} and \emph{l2 norm scale}.
For this study, we sample \SI{10}{\percent} stratified from the training set to serve as a validation set.
All parameters were chosen from a uniformly random discrete or continuous distribution.
The ranges and the parameters yielding the best performance on the validation set are given in \Cref{table:clstmsearch}.

\begin{table}
    \centering
    \small
    \begin{tabular}{l  c  c c}
        \toprule
        parameter & min & max & selected \\
        \midrule
        number of filters & 10 & 500 & 384 \\
        filter width & 2 & 5 & 3 \\
        rnn cell units & 16 & 500 & 93 \\
        dropout rate & 0.0 & 0.5 & 0.23 \\
        l2 normalization scale & 0.0 & 3.0 & 0.79 \\
        \bottomrule
    \end{tabular}
    \caption{\clstm parameters and settings selected by random search from search ranges of [\emph{min}, \emph{max}].}
    \label{table:clstmsearch}
\end{table}

Using the determined optimal parameter settings, models with both types of embeddings were trained on the full training
set and evaluated on the test set.
\Cref{table:clstmresults} shows that the \clstm model performs well on the scientific embeddings, but consistently worse
than the SVM model using handcrafted features and achieves a macro-F1 score of $67.49$ and $67.02$ for subtask 1.1 and
subtask 1.2 respectively.

\begin{table}
    \centering
    \small
    \begin{tabular}{ l c c c c}
        \toprule
         & \multicolumn{2}{c}{subtask 1.1}  &  \multicolumn{2}{c}{subtask 1.2}
         \\
         & macro F1 & micro F1 & macro F1 & micro F1 \\
         \midrule
        \ccm & 54.42 & 67.61 & \textbf{74.42} & \textbf{78.87} \\
        \scm & \textbf{67.49} & \textbf{70.96} & 67.02 & 74.37 \\
        \bottomrule
    \end{tabular}
    \caption{Results for \clstm models trained with \ccm and \scm embeddings on both subtasks.}
    \label{table:clstmresults}
\end{table}

    \section{Discussion}
\label{sec:discussion}
We briefly discuss our approach during the training phase of the SemEval-Challenge and how label distribution and
evaluation measure influences our results.
Ahead of the final evaluation phase where the concealed test data $D_{test}$ was presented to the participants we were
given a preliminary test partition $D_{pre}$ as part of the training data $D_{train}$.
To be able to estimate our performance we evaluated it on $D_{pre}$ as well as for a 10-fold stratified
cross validation setting.
We chose this procedure to be sure to pick the best system for submission at the challenge.

As some classes were strongly underrepresented in the training corpus and $D_{pre}$, we assumed that this is also true
for the final test set $D_{test}$.
When in doubt we therefore chose to optimize according to $D_{pre}$ as cross validation is based on a
slightly more balanced data set (of train data for subtask 1.1 + 1.2).
The best system we submitted for subtask 1.1 of the challenge achieved a macro-F1 of 75.05\% on $D_{pre}$ during the
training phase which shows that we were able to estimate our final result pretty closely.

During training we also noticed that for heavily skewed class distributions as in this case, macro-F1 as an evaluation
measure strongly depends on a good prediction of very small classes.
For example, macro-F1 of subtask 1.1 increases by 5 points if we correctly predict one \texttt{Topic} instance out of
three instead of none.
Thus we pick a configuration that optimizes the small classes.

We also omitted some lexical feature sets from our system as performance on the temporary and final test set showed
that they did not improve results.
These features were hypernyms of context and entity tokens from WordNet and dependency paths between entities.
Using tf-idf normalization instead of boolean for lexical features also worsened our results.

The \clstm model performes quite well, considering it only relies on very limited information, the sequence of entity
and word embedding vectors, to perform the classification.
For example the model has no way of determining the direction of the relation and we speculate that increasing the
model complexity to include such information might increase the performance further.
Additionally, the results for subtask 1.2 show that in contrast to the SVM model, \clstm does not perform consistently
better with \scm embeddings, which warrants further investigation.

    \section{Conclusion}\label{sec:conclusion}
In this paper, we described our SemEval-2018 Task 7 system to classify semantic
relations in scientific literature for clean (subtask 1.1) and noisy (subtask 1.2)
data and its results during the post-evaluation phase.
We constructed features based on relation entities and their context by means of hand-crafted lexical features as
well as word embeddings.
To better adapt to the scientific domain, we trained scientific word embeddings on a large corpus of scientific papers
obtained from arXiv.org.
We used an SVM to classify relations and additionally contrasted these results with those obtained from training a
C-LSTM model on the scientific embeddings.
Due to improved preprocessing we were able to obtain a macro-F1 score of \SI{75.11}{\percent} on clean data and
\SI{81.44}{\percent} on noisy data.
We finished the challenge as 4th out of 28 (subtask 1.1) and 6th out of 20 (subtask 1.2) though the results
from~\citet{hettinger2018semeval} are applied.

In future work, we will improve the tokenization of the scientific word
embeddings and also take noun compounds into account, as they make up a large
part of the scientific vocabulary.
We will also investigate more complex neural network based models, that can leverage additional information, for
example relation direction and POS tags.

    \bibliography{ms}
    \bibliographystyle{resources/acl_natbib}

\end{document}